\documentclass[runningheads]{llncs}
\usepackage{graphicx}
\usepackage{algorithm}
\usepackage{algpseudocode}
\usepackage{amsmath}
\usepackage{amssymb}
\usepackage{booktabs}
\usepackage{hyperref}
\usepackage{array}
\usepackage{multirow}
\usepackage{booktabs} 
\usepackage{tabularx}
\usepackage{adjustbox}
\usepackage{subcaption}
\usepackage{float}
\usepackage{amsfonts} % or \usepackage{amssymb}

\newcommand{\sgn}{\operatorname{sgn}}
\begin{document}
\title{Enhancing Counterfactual Explanation Search with Diffusion Distance and Directional Coherence}
\author{Marharyta Domnich\inst{1}\orcidID{0000-0001-5414-6089} \and
Raul Vicente\inst{1}\orcidID{0000-0002-2497-0007}}
\authorrunning{Domnich et al.}
\titlerunning{Enhancing Counterfactual Explanation Search}
\institute{Institute of Computer Science, University of Tartu, Tartu, Estonia 
\email{marharyta.domnich@ut.ee}\\
}
\maketitle              % typeset the header of the contribution
\begin{abstract}
A pressing issue in the adoption of AI models is the increasing demand for more human-centric explanations of their predictions. To advance towards more human-centric explanations, understanding how humans produce and select explanations has been beneficial. In this work, inspired by insights of human cognition we propose and test the incorporation of two novel biases to enhance the search for effective counterfactual explanations. Central to our methodology is the application of diffusion distance, which emphasizes data connectivity and actionability in the search for feasible counterfactual explanations. In particular, diffusion distance effectively weights more those points that are more interconnected by numerous short-length paths. This approach brings closely connected points nearer to each other, identifying a feasible path between them. We also introduce a directional coherence term that allows the expression of a preference for the alignment between the joint and marginal directional changes in feature space to reach a counterfactual. This term enables the generation of counterfactual explanations that align with a set of marginal predictions based on expectations of how the outcome of the model varies by changing one feature at a time. We evaluate our method, named Coherent Directional Counterfactual Explainer (CoDiCE), and  the impact of the two novel biases against existing methods such as DiCE, FACE, Prototypes, and Growing Spheres. Through a series of ablation experiments on both synthetic and real datasets with continuous and mixed-type features, we demonstrate the effectiveness of our method.

\end{abstract}
\section{Introduction}
The demand for explainability of AI models has reached a new level of urgency with the rapid adoption of AI in different domains. Fueled by deep learning algorithms, applications in critical areas such as medical imaging \cite{miotto2018deep,shen2017deep}, law \cite{chalkidis2019deep}, or finance \cite{heaton2017deep} are seeking to automate and assist in decision-making \cite{whittaker2018ai}. However, the ability to explain why a certain prediction or decision was inferred is a fundamental prerequisite for responsible deployment in high-stake fields where accountability, transparency, and trust are valued \cite{schmidt2020transparency,dignum2019responsible}.

Developing explanations of AI models for humans confront us with the complexity of human explanatory processes \cite{kulesza2013too,hilton1996mental}.  Moreover, humans excel at recognizing patterns from limited examples, even in uncertain situations, and can generalize concepts to address new problems \cite{holzinger2023toward}. While producing and evaluating explanations is natural to us, the underlying processes depend on complex mental models of the world which assist in inferring meaning from incomplete information based on previous experience \cite{keil_explanation_2006}. Ideally, accessing a human’s world model would allow for producing explanations that fill the specific gaps in understanding, by contrasting input with prior knowledge of one’s mental model. Without access to an accurate “theory of mind” model \cite{aru2023mind} as part of one’s explainable model, another pathway to more human-centric explanations is to incorporate the preferences or biases that humans have reported in their judgment of explanations \cite{sokol2020explainability}. Following this perspective, our work %introduces a novel framework for %generating counterfactual explanations, CoDiCe, which 
incorporates two cognitive biases in a novel way: feasibility with diffusion distance and directional coherence as consistency of proposed changes with prediction direction.

Counterfactual explanations have emerged as a powerful tool in Explainable AI (XAI). Research in social sciences \cite{miller_explanation_2019} highlights that human explanations are usually contrastive, presented in a format “X because of Y rather than Z”, where the “rather than Z” part is not always explicitly mentioned. A significant subtype of contrastive explanations are counterfactual explanations which answer questions such as “What would have happened if X had not occurred?”. In the context of machine learning model explainability, counterfactual explanations are used as local explanations \cite{karimi2022survey,guidotti_counterfactual_2022}. Given a particular instance and a trained model, they answer the question “What should the input have been in order to change the decision outcome?”. The power of counterfactual explanations lies in their ability to present alternative realities, aligning closely with our natural hypothetical reasoning \cite{zemla2017evaluating}. 

Despite significant advancements in the development of counterfactual methods such as DiCE \cite{mothilal_explaining_2020}, FACE \cite{poyiadzi_face_2020}, DACE \cite{kanamori_dace_2020}, CLUE \cite{antoran_getting_2020}, Guided Prototypes \cite{van2021interpretable}, CARE \cite{rasouli2024care} and others, a complete integration of coherence within these approaches continues to be challenging. There is a consensus within the community that coherence is a crucial aspect distinguishing a basic explanation from an effective one. However, the multifaceted nature of coherence complicates its definition and application within AI explanations. Zemla et al. \cite{zemla2017evaluating} highlight this complexity by distinguishing between internal and external coherence, where the internal coherence relates to the consistency among the components of an explanation, while external coherence concerns the explanation's alignment with the user's pre-existing knowledge. While existing approaches have attempted to model external coherence through user-defined constraints or by tracing the directional correlations within the feature space, the influence of the marginal direction of features concerning model predictions has not been explored. On top of that, the use of diffusion distance ensures that counterfactual points are obtained through feasible paths of connected points respecting underlying geometry of the data manifold.  
%Despite progress in developing various counterfactual methods, such as DiCE\cite{mothilal_explaining_2020}, FACE\cite{poyiadzi_face_2020}, DACE\cite{kanamori_dace_2020}, CLUE\cite{antoran_getting_2020}, there is still no agreement on what makes a good explanation. However, there is yet little consensus on the desirable properties of a counterfactual explanation [D.Slack ”Counterfactual Explanations Can be Manipulated” \cite{slack2021counterfactual}, “Issues with post-hoc CF Explanations”\cite{laugel2019issues}, Guidotti “Counterfactual Explanations and how to find them” \cite{guidotti_counterfactual_2022}, Karimi “Algorithmic recourse: from counterfactual explanations to interventions” \cite{karimi2022survey}]. Many approaches study the problem only from a geometrical viewpoint, potentially overlooking a crucial dimension of explanation effectiveness, which is to be aligned with human cognitive processes \cite{byrne2023good}[Ruth M.J. Byrne]. 

Our work contributes to this field by introducing CoDiCE, a framework designed to generate Directionally Coherent Counterfactual Explanations. This approach diverges from traditional methods by:

\begin{enumerate}
    \item Replacing standard $L_p$ distance measures used for proximity with diffusion distance, prioritizing connectivity and feasibility of transitions towards counterfactual scenarios. 
    \item Incorporating directional coherence as a constraint. This guides the selection of counterfactual explanations which joint changes (changing multiple features simultaneously) aligned with desired marginal changes (model of how outcome should vary if one changes one feature at a time).
\end{enumerate}

\section{Related Work}
Our work builds upon the literature about counterfactual explanations, by focusing on their interpretation and integration of two cognitive biases: feasibility and coherence.

To account for \textbf{feasibility}, different notions of distances between the original point and counterfactual point were proposed. An early approach by Wachter et al. \cite{wachter2017counterfactual} posits Manhattan distance, adjusted by the inverse median absolute deviation, as a measure. Alternatives to this include the Euclidean (L2) or Gower distances, with some methods employing a blend of L1 and L2 distances weighted variably as an elastic net penalty. Despite accounting for scale variability, these metrics might neglect data density variations, potentially placing the counterfactual outside the data manifold. To mitigate this, CEM \cite{dhurandhar2018explanations} propose training an auto-encoder on the desired class data, introducing a novel objective function term that penalizes the deviation of a counterfactual from its auto-encoded representation. Similarly, \cite{van2021interpretable} expands upon this concept by identifying a prototype or class-representative instance through the autoencoder's latent space. Furthermore, the DACE method \cite{kanamori_dace_2020} utilizes Mahalanobis distance, which accounts for data correlations and Local Outlier Factor that penalize points that are out of distribution.

Nonetheless, these methodologies do not sufficiently consider the transition path from the original instance to its counterfactual. In contrast, FACE \cite{poyiadzi_face_2020} leverages k-NN to construct a connectivity graph, subsequently employing Dijkstra’s algorithm to identify the most feasible counterfactual pathway. This strategy not only affirms the feasibility of the counterfactual point but also of its pathway. A notable limitation of the FACE algorithm is its reliance on endogenous data points, which can complicate feasibility in sparse datasets or higher-dimensional spaces. Inspired by FACE, our approach utilizes diffusion distance, initially training a diffusion map to generate a transition graph within the diffusion space. This strategy accounts for data flow, enabling the projection of new points onto diffusion coordinates without the constraints imposed by reliance on endogenous data.

An explanation is deemed \textbf{coherent} if it aligns with the recipient's existing beliefs and knowledge, essentially reflecting the user's mental model of the application domain. Echoing Zemla et al. \cite{zemla2017evaluating} understanding that there are two types of coherence, Forster et al. \cite{forster2023user} suggest that an explanation gains coherence from consistency with the user's knowledge. Additionally, when the counterfactual scenario depicted is both realistic and typical of the alternative class distribution, they incorporate a loss term based on density estimate and external knowledge, adding pre-defined constraints into the search. Alternatively, CARE framework \cite{rasouli2024care} interprets coherence as the consistency between the altered and unaltered features from the original to the counterfactual point by training model of correlations which guide the search towards more correlated features. 
Various methods interpret the coherence measure as ensuring the point remains within the same class distribution, modifying proximity with the Mahalanobis distance \cite{cheng_dece_2021}, or incorporating auto-encoders \cite{dhurandhar2018explanations,van2021interpretable} to maintain the counterfactual within the data manifold. 

Beyond data distribution conformity, coherence involves the transition from the original point to the plausible counterfactual point. Some methods, such as that proposed by \cite{mahajan2019preserving}, advocate for incorporating partial causal knowledge into the search process, suggesting learning feasibility constraints from user feedback. This approach, while emphasizing feasibility, indirectly fosters coherence by ensuring transitions adhere to plausible causal relationships. Additionally, Raman et al. \cite{raman2023bayesian} utilize a Bayesian approach to model the relationships between variables using conditional distributions. This allows for sampling counterfactuals from the posterior density while preserving domain-specific constraints. Primary methods underscore the significance of coherence by ensuring counterfactual explantions adhere to partial domain constraints. However, most approaches overlook the coherence of transitions from factual to counterfactual points with respect to the model's output.

%Sparsity story if we have to:
%From another perspective [Miller] one desirable property is simplicity. However, some findings [Ruth] report that explanations should appeal to a good number of causal mechanisms in order to be trustworthy. In counterfactual explanation research simplicity is approximated with sparsity. Sparsity ensures explanations are concise and focus on a minimal set of changes. Sparsity is usually incorporated as part of distance either Manhattan or elastic net. Another way with Minimal Feature Boundary paper, where they take a subset of features that brings maximum mutual information gain.
%Diversity story:   
%[DiCE] \cite{mothilal_explaining_2020}, Certify, CARE

%Robustness:  \cite{virgolin2023robustness}
%Counterfactual input should be robust meaning that small changes which are not controlled by user should not change the output decision. The importance of that was highlighted in \cite{virgolin_robustness}, \cite{artelt_evaluating_2021}.  Alternatively, counterfactual change should not be located in region of high uncertainty, since the outcome of that region is not stable \cite{le_improving_2022}, \cite{slack_reliable_2021}.

\section{Incorporating novel biases in counterfactual search}
This section introduces our approach to refining the search for counterfactual explanations by incorporating two terms that account for a feasibility and external coherence biases in a novel way. The methodology aims to generate more intuitive and human-centric explanations. Finally, we detail the integration of these biases into the counterfactual objective function and optimization strategy. 

\subsection{Using diffusion distance to search for more feasible transitions}

\begin{figure}[ht]
    \centering
    \begin{subfigure}[b]{0.63\linewidth}
        \includegraphics[width=\linewidth]{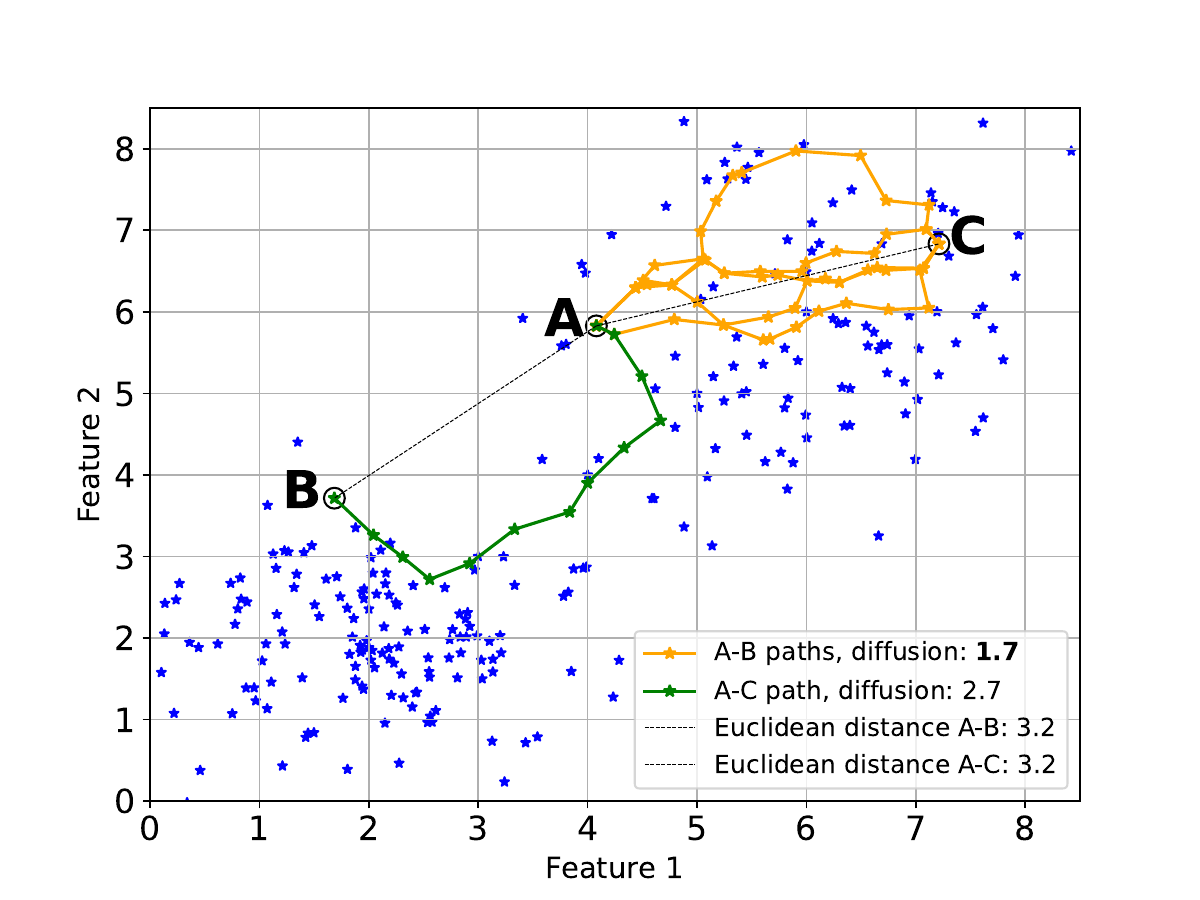}
        %\caption{Comparison of Euclidean distance and diffusion distance between A-B and A-C. While Euclidean distance is the same, there are more trajectories between A and C than between A and B, hence the diffusion distance $AC < AB$}
        \label{fig:first-diff}
    \end{subfigure}
    \hfill % Optional spacing
    \begin{subfigure}[b]{0.33\linewidth}
        \includegraphics[width=\linewidth]{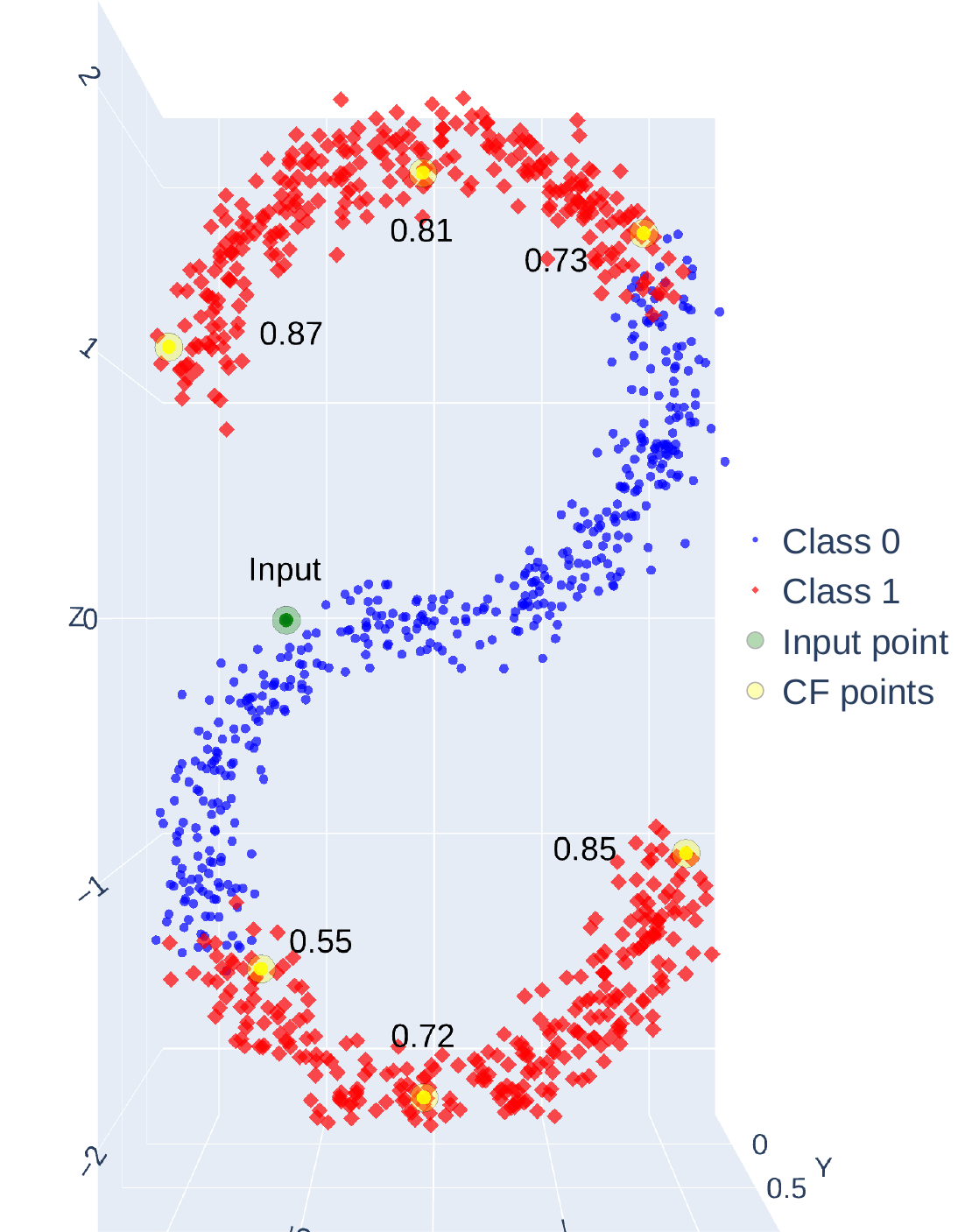}
        %\caption{Synthetic dataset with two classes. The input point belongs to class 0, the diffusion distances between that point and 6 counterfactual candidates are displayed.}
        \label{fig:second-diff}
    \end{subfigure}
    \caption{Illustration of the concept of diffusion distance and its use for counterfactual search. Left panel: Points connected by numerous short distance paths (A-C) exhibit a shorter diffusion distance than pairs of points which connections pass through a bottleneck or low density region (A-B). Note that evaluated by Euclidean distance the pairwise distance A-C and A-B would be exactly the same. Right panel: 3D S-shaped synthetic dataset with two classes. The input point belongs to class 0, the diffusion distances between such point and 6 counterfactual candidates are displayed.}
    \label{fig:both-images}
\end{figure}

A strategy for generating meaningful counterfactual explanations is to develop methodology that emphasizes the feasibility, coherence, and actionability of possible explanations. We propose the utilization of diffusion distance as a  metric to assess the connectivity and potential actionability of counterfactual transitions. Unlike traditional distance metrics such as Euclidean (L2), Manhattan (L1), or shortest-path distance on the data manifold (as used in FACE or Isomap \cite{tenenbaum2000global}), diffusion distance offers a nuanced understanding of the data manifold by prioritizing transitions between data points that are interconnected through numerous, short paths. This approach brings points that are highly connected by numerous short paths into closer proximity, and hence highlighting points for which numerous short routes exist to transition from one point to the other while being on the data manifold. The concept of diffusion distance and its role in detecting counterfactual points that are more "accessible" from the original instance (in the sense of the existence of numerous short distance routes between the points) is illustrated in Fig.~\ref{fig:both-images}.   

%but also illuminates actionable pathways for counterfactual transitions, as illustrated in Fig.~\ref{fig:both-images}.

The formal definition of diffusion distance between two points $x$ and $y$ at time $t$ is given by:

\begin{equation}
D_{\text{diff}}(x, y, t)^{2} = \sum_{z} \frac{(p_t(x| z) - p_t(y| z))^2}{\phi_0(z)} \, ,
\end{equation}

\noindent where $p_t(x| z)$ represents the probability of transitioning from point $z$ to $x$ in $t$ steps following a diffusion process (random walk on the graph), and $\phi_0(z)$ is the stationary distribution of the diffusion process at point $z$. This formula highlights the diffusion distance's capacity to account for the data's intrinsic geometry through probabilistic transitions.

\subsubsection{Employing Diffusion Distance with Self-Tuning Kernel for Local Scaling.}
Our implementation incorporates a self-tuning kernel within the diffusion distance framework \cite{zelnik2004self}, adjusting dynamically to the variance in the data. This adjustment ensures the robustness across different data domains and reduce the number of parameters needed for fitting, as the only parameter we ask is number of nearest neighbors, which is intuitive to set up. 

Diffusion distance is rooted in the concept of diffusion processes on graphs, encapsulating the connectivity and density of data points within a dataset. This metric quantifies the ease of traversing the data landscape from one point to another, factoring in the multitude of potential paths and their associated probabilities. The key advantage of diffusion distance over the shortest-path distance on the manifold as used by FACE or Isomap is its robustness to noise, which is particularly valuable in high-dimensional settings where data sparsity and noise are prevalent challenges. By facilitating the exploration of multiple paths, diffusion distance looks for counterfactuals through a sequence of realistic transitions which favour to in-distribution feasible point.

\subsection{Directional coherence}
Directional Coherence formulates a bias designed to maintain consistency between the marginal (one feature at a time) and joint (multiple features simultaneously) directions in feature space needed to flip the outcome of the model's prediction. This coherence facilitates the generation of counterfactual explanations that not only adhere to the model's predictions for individual feature alterations but also align with the overall direction of change necessary to shift to a desired counterfactual state. Such term can be use to tune the importance of aligning counterfactual paths with intuitive human reasoning about a set of causal expectations when changes are produced in marginal directions (changing one feature at a time). 

\begin{figure}
\includegraphics[width=\textwidth]{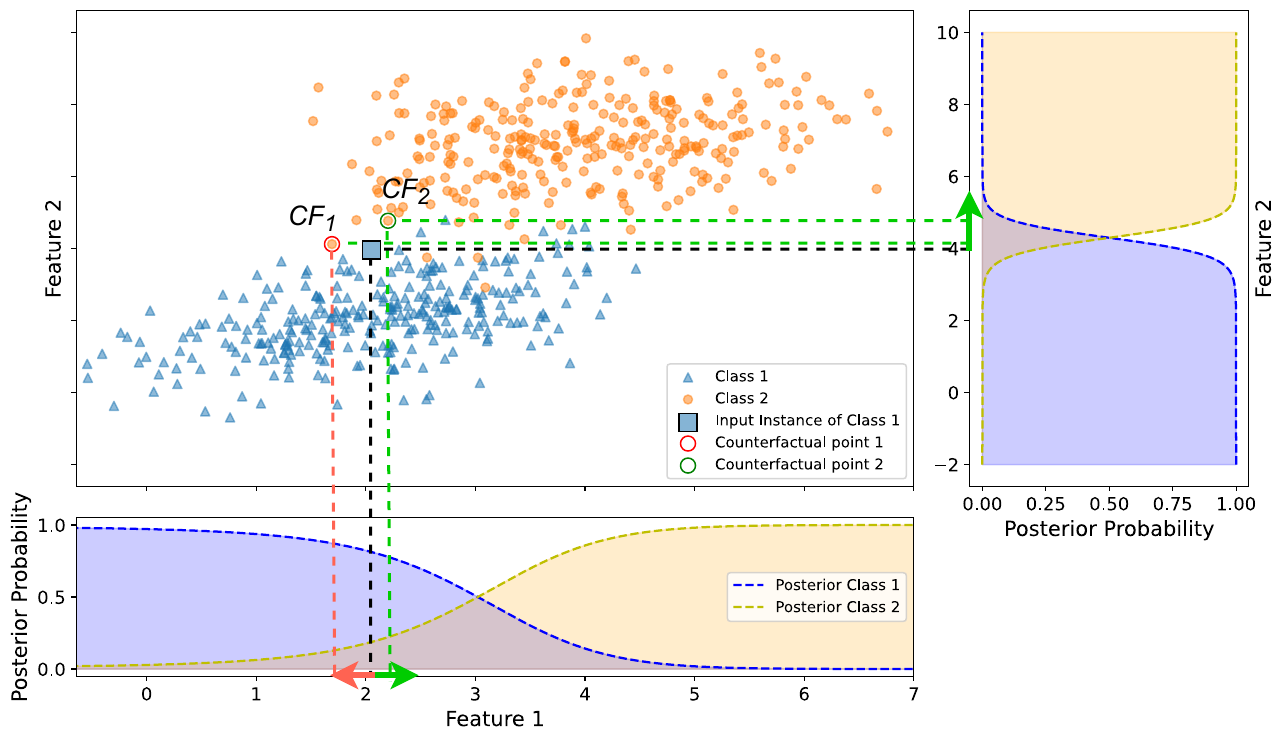}
\caption{Illustration of Directional Coherence. The input point belongs to Class 1. Given counterfactual candidates $CF_1$ and $CF_2$ at equal distance from the original input point, we deem $CF_1$ as incoherent with respect to the expected effect of changing Feature 1. Intuitively, $CF_1$ suggests to decrease Feature 1, while the effect of increasing either Feature 1 or Feature 2 is to increase the posterior probability of predicting Class 2. For the other counterfactual ($CF_2$), there is an agreement between the direction of marginal changes (changing one feature at a time) and the joint direction of changes resulting in a more coherent counterfactual point.} 
\label{fig:directional_coherence}
\end{figure}

To illustrate this concept, consider the scenario of applying for a home loan, where it is intuitively expected that an increase in income for either the applicant or co-applicant would improve the chances of loan approval. We would be shocked to learn that a bank advises increasing  the applicant's income, but decrease a co-applicant income. This counterintuitive recommendation could arise from the specific nature of the data distribution, reflecting scenarios where other people with  these factual scenario in the past got loan approval. We argue that although observing such point is possible, it would represent an undesirable direction for counterfactual explanation. Figure~\ref{fig:directional_coherence} illustrates this conceptual situation. An input point highlighted with rectangular shape and belonging to Class 1, has two counterfactual candidates $CF_1$ and $CF_2$, which are associated with the desired Class 2. The data spread indicates that increases in Feature 1 and Feature 2 are correlated with a higher likelihood of predicting Class 2. Consequently, $CF_2$ point is directionally coherent, as the joint increase in these features aligns with the marginal direction of probability of Class 2. On the other hand, $CF_1$ is directionally incoherent, since the change in Feature 1 leads to decrease in the posterior probability of predicting Class 2.

Thus, directional coherence is predicated on the intuition that certain feature alterations should consistently lead to predictable changes in the model's output. This is especially crucial in complex domains where interpretability and actionability of counterfactual explanations are required. By assessing the directional impact of each feature independently, we can ascertain the collective influence exerted by all features on the transition towards the desired counterfactual state.

Mathematically, we formulate directional coherence as a term that quantifies the preference for alignment between joint and marginal directional changes in the feature space necessary to achieve a counterfactual outcome. For clarity, we introduce here the case of a classifier. The corresponding formulation for a regression model is an straightforward extension.

Let us denote $f: \mathcal{X} \rightarrow \mathcal{Y}$ the classification model.
Given an original instance as a vector $x=(x_1, x_2,\dots, x_n) \in \mathcal{X} \subseteq \mathbb{R}^n$ and a counterfactual instance $x^\ast=(x^\ast_1, x^\ast_2,\dots, x^\ast_n) \in \mathcal{X} \subseteq \mathbb{R}^n$ that brings the desired outcome. The goal is to evaluate the coherence of the transition from $x$ to $x^\ast$ in achieving a specified outcome label $y$ with a set of expected marginal transitions $\{x_{i} \rightarrow x'_{i} \,\, | \,\, f(y|x_{1},x_{2},\ldots,x'_{i},\ldots,x_{n}) \geq f(y|x_{1},x_{2},\ldots,x_{i},\ldots,x_{n}) \, , 1 \leq i \leq n \}$. Notably, while marginal transitions are typically derived from the model, user-specified marginal transitions, when provided, take precedence over those suggested by the model.  

Then, the Directional Coherence score counts the excess of features which have aligned marginal ($'$) and joint ($\ast$) directions to increase the model's prediction probability towards the desired outcome y:

\begin{equation}
dcoherence = \frac{1}{n}\sum_{i=1}^{n}\sgn\left(\left(x^\ast_i-x_i\right)\left(x'_i-x_i\right)\right) \, .
\end{equation}

The information about incoherent features can be leveraged to introduce new constraints or refine the model. Additionally, repeated patterns of incoherence could indicate areas where the model's sensitivity to changes in feature values needs further investigation or adjustment. The implementation of calculating directional coherence is illustrated in Appendix A Algorithm~\ref{alg:dircoh}.

\subsection{Bringing feasibility and directional coherence into counterfactual objective function}

%Incorporating two aforementioned biases significantly enhances the counterfactual search. 
Next we formalize the incorporation of the feasibility and coherence biases within the objective function for the counterfactual search. We denote by $f$ a trained predictor function that maps the input space $\mathcal{X}$ to the output space $\mathcal{Y}$, i.e., $f: \mathcal{X} \rightarrow \mathcal{Y}$. Given a factual point or the original input point $x=(x_1, x_2,\dots, x_n) \in \mathcal{X} \subseteq \mathbb{R}^n$, our objective is to identify a counterfactual point $c^\ast=(c^\ast_1, c^\ast_2,\dots, c^\ast_n) \in \mathcal{X} \subseteq \mathbb{R}^n$ that yields the desired label $y$ while minimising a weighted sum of diffusion distance, sparsity, and directed coherence penalties. The optimization problem is defined as follows:
\begin{align}
c &= \arg \min_{x^\ast} \Big( \text{loss}(f(x^\ast), y) + \lambda_1 \text{diffusion\_dist}(x^\ast, x) \nonumber \\
&\quad + \lambda_2 \text{sparsity}(x^\ast, x) + \lambda_3(1-\text{dcoherence}(x^\ast, x)) \Big) \, ,
\end{align}
\noindent where:
\begin{itemize}
    \item $\text{loss}(f(x^\ast),y)$ is the loss term that checks if the counterfactual outcome is equal to the desired outcome, we utilize commonly used loss measures hinge-loss \cite{gentile1998linear} for classification and mean squared error \cite{prasad1990estimation} for regression. 
    \item $\text{diffusion\_dist}(x^\ast, x)$ quantifies the diffusion distance between the original point $x$ and the counterfactual point $x^\ast$ (see formula (1)).
    \item $\text{sparsity}(x^\ast, x)$ computes the $l_0$ distance to count the number of features that have been modified.
    \item $\text{dcoherence}(x^\ast, x)$ assesses the directional coherence by aligning the joint direction of the counterfactual point with its marginals. Since we are interested in minimizing objective function, we take the penalty measure $(1-\text{dcoherence})$.
\end{itemize}
The terms are weighted by hyperparameters $\lambda_1$, $\lambda_2$, and $\lambda_3$, which can be adjusted or set to $0$ if a particular constraint is not applicable.

There is a natural division between continuous and categorical features which have different properties. For \textbf{categorical features}, the notion of diffusion distance does not apply, and measuring distance is challenging, necessitating a context-specific approach. We employ the $L_0$ norm to detect category changes, which could be further refined based on the number of categories and the difficulty of changing each feature.:
\begin{align}
dist_{cat}(x, x^\ast) = \frac{1}{m}\sum_{j=1}^{m}I(x^{\ast}_j \neq x_j)
\end{align}
It is notable that the distance for categorical features overlap with the sparsity term, we intend to keep it that way, as the user might want to tune down sparsity weight term without accounting for its effect on the search of counterfactual explanation. 
Therefore, the objective function for counterfactual search for mix-type data is the following:
\begin{align}
c &= \arg \min_{x^\ast} \Big( \text{loss}(f(x^\ast), y) + \lambda_1 \text{diffusion\_dist}_{cont}(x^\ast, x) + \lambda_1 dist_{cat}(x^\ast, x) \nonumber \\
&\quad + \lambda_2 \text{sparsity}(x^\ast, x) + \lambda_3(1-\text{dcoherence}(x^\ast, x)) \Big)
\end{align}

We optimize this expression with genetic algorithm, similarly to DiCE 'genetic' optimization \cite{mothilal_explaining_2020} with the details of implementation described in Appendix A Algorithm~\ref{alg:ga_counterfactual}.

\subsection{Evaluation metrics}
To assess the quality and compare the performance of the generated counterfactual explanations, we utilize commonly accepted metrics (Validity, Weighted L1 Continuous, Categorical L0) as well us our novel metrics inspired by our integrated biases: Diffusion Distance and Directional Coherence. For their detailed description see Appendix A. 

\section{Experiments}
We conducted expereiments on two synthetically generated datasets to visualize the effect of diffusion distance. After that, we applied our framework for commonly used in counterfactual explanation literature classification datasets, such as Diabetes, Breast Cancer  \cite{misc_breast_cancer_wisconsin_(diagnostic)_17} and
mixed-type features Adult \cite{misc_adult_2} and German \cite{misc_statlog_(german_credit_data)_144}.
Additionally, we used energy consumption prediction dataset \cite{sakkas2023explainable} for testing regression settings as for this case we also had access to domain experts for feedback.
\subsection{Synthetic datasets}
To illustrate the effect of diffusion distance we generated two synthetic datasets: an S surface and a Swiss roll, utilizing the sklearn.datasets module for their creation. We thresholded the parameter of the generated shapes, dividing each dataset into two distinct classes. The dataset were partitioned into training and test subsets with a 80/20 split. We trained a Support Vector Machine (SVM) classifier with radial basis kernel on these datasets resulting in 97\% accuracy on the test set for a S surface and 99\% accuracy for a Swiss roll.

\subsection{Datasets with continuous features}

\textbf{The Diabetes dataset}, sourced from the UCI Machine Learning Repository, consists of diagnostic measurements for predicting the onset of diabetes within a Pima Indian population. It features 768 instances, each with 8 numeric predictor variables such as the number of pregnancies, plasma glucose concentration, blood pressure, and body mass index, among others. The outcome variable is binary, indicating the presence or absence of diabetes.

\textbf{The Breast Cancer Wisconsin (Diagnostic) dataset}, also from the UCI Machine Learning Repository, comprises features computed from a digitized image of a fine needle aspirate (FNA) of a breast mass. It includes 569 instances with 30 continuous features, describing characteristics of the cell nuclei present in the image. The prediction task is binary, distinguishing between malignant and benign tumors.

\textbf{Energy consumption dataset} is a time series data of residential building energy consumption contains of 473 instances with indoor temperature, outdoor temperature, active electricity historical observations. The prediction feature is future electricity consumption. The counterfactual question is what should have been the input to decrease energy consumption. Details of the dataset are available here \cite{sakkas2023explainable}.

\subsection{Classification datasets with mix-type features}

\textbf{The Adult Income dataset}, often referred to as the "Census Income" dataset, contains demographic information from the 1994 Census database. It consists of 48842 instances and 14 features (6 numerical and 8 categorical), including age, work class, education, marital status, occupation, and hours per week, among others. The dataset's target variable is binary, predicting whether an individual's income exceeds \$50K/year.

\textbf{The German Credit} dataset comprises financial and demographic data for 1,000 loan applicants. Each instance is described by 20 attributes, a mix of continuous and categorical variables, such as credit history, savings account balance, employment duration, and purpose of the loan. The objective is to classify individuals into good or bad credit risks.

\textbf{Compas} dataset encompasses information related to defendants involved in the criminal justice system. It consists of 7214 instances with 11 features (4 continuous and 7 categorical), such as age, race, gender, criminal history, charge degree, etc. The target variable is the risk of recidivism (high or low). The dataset has been a point of analysis in discussions on the fairness, bias, and transparency of predictive algorithms in legal settings. 

\subsection{Benchmarking with other frameworks}
We compared our results with various counterfactual frameworks, focusing on validity, diffusion distance, weighted L1 distance, directional coherence, and L0 categorical for mixed-type data. The compared methods are:

\textbf{Diverse Counterfactual Explanations (DiCE)} \cite{mothilal_explaining_2020}, the most popular method in literature, generates diverse counterfactual instances using weighted L1 for proximity on continuous features and L0 for categorical. Acknowledging the trade-off between diversity and proximity, we generated a single, optimal counterfactual explanation using DiCE's original implementation.

\textbf{Feasible and Actionable Counterfactual Explanations (FACE)} \cite{poyiadzi_face_2020} considers feasible paths using the Dijkstra algorithm and produces counterfactual points from existing training data. We ran the FACE algorithm using the CARLA benchmark library \cite{pawelczyk_carla_2021} for algorithm comparison.

\textbf{Guided Prototypes (Prototypes)} \cite{van2021interpretable} integrates the notion of coherence by training an auto-encoder to select a prototype instance, ensuring the typicality of the point. Prototypes use a combination of weighted L1 and L2 as an elastic net regularizer for proximity.

\textbf{Growing Spheres (GS)} \cite{laugel2018comparison} employs weighted $L_2$ and $L_0$ norms between counterfactuals and candidates. It is often reported to generate the closest counterfactuals in terms of proximity. Importantly, GS does not support mixed-feature datasets.

We ran Prototypes and GS methods from \cite{moreira2022benchmarking}, a library created for benchmarking counterfactual methods.

\section{Results}

\subsection{Diffusion distance and Directional coherence on synthetic and Diabetes datasets}
\begin{figure}[!htbp]
    \centering
    % First row of two subfigures
    \begin{subfigure}[b]{0.48\linewidth}
        \includegraphics[width=\linewidth]{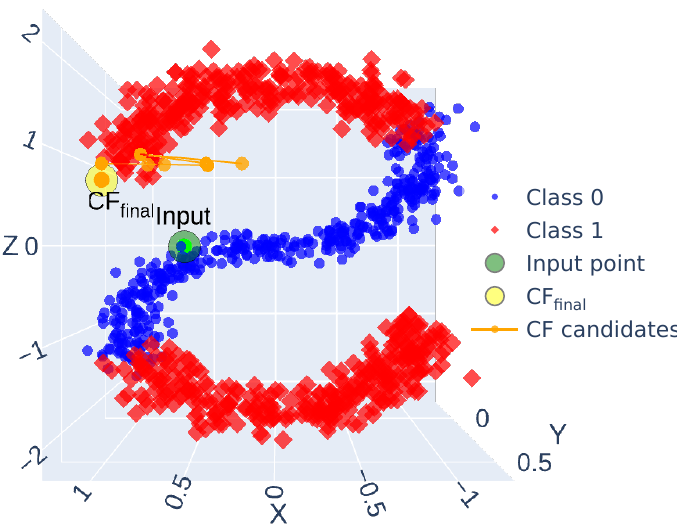}
        \caption{}
        \label{fig:first}
    \end{subfigure}
    \hfill % Space between the first and second subfigures
    \begin{subfigure}[b]{0.48\linewidth}
        \includegraphics[width=\linewidth]{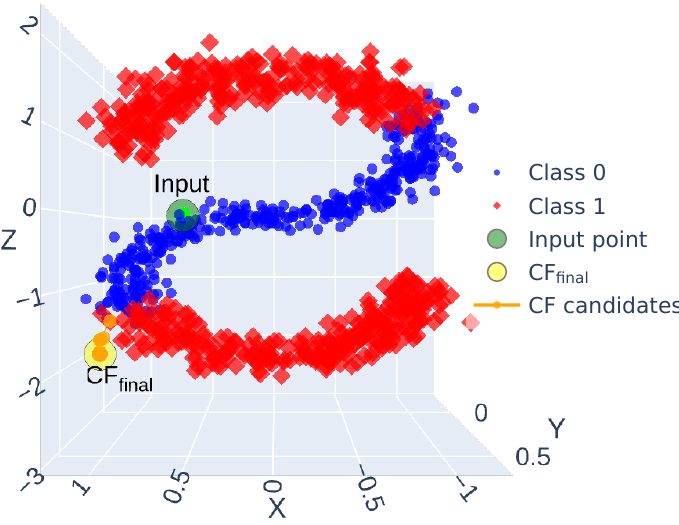}
        \caption{}
        \label{fig:second}
    \end{subfigure}
    % Add some vertical space between the rows
    %\vspace{1em} % Adjust the vertical space as needed
    
    % Second row of two subfigures
    \begin{subfigure}[b]{0.48\linewidth}
        \includegraphics[width=\linewidth]{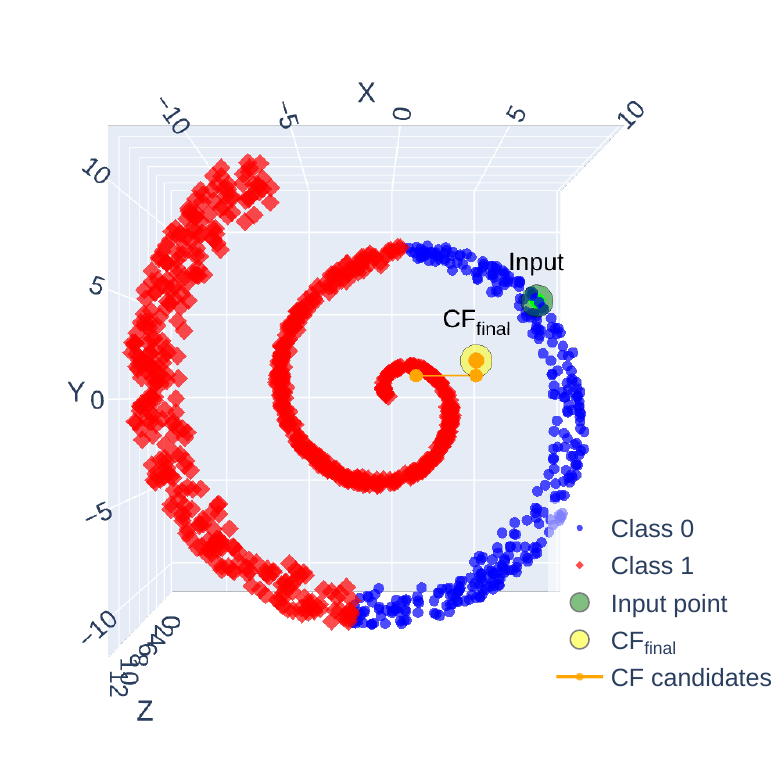}
        \caption{}
        \label{fig:third}
    \end{subfigure}
    \hfill % Space between the third and fourth subfigures
    \begin{subfigure}[b]{0.48\linewidth}
        \includegraphics[width=\linewidth]{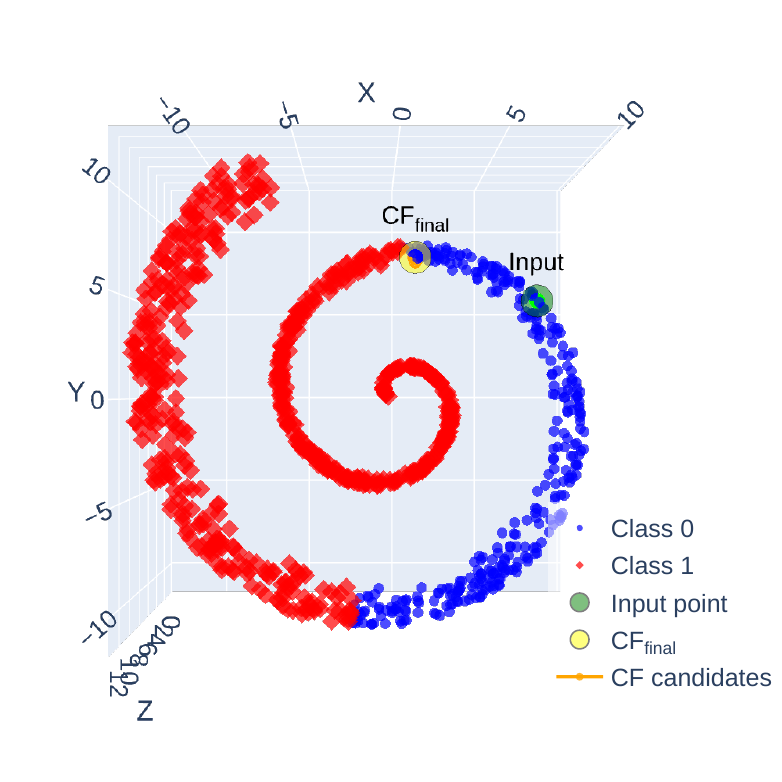}
        \caption{}
        \label{fig:fourth}
    \end{subfigure}
    
    \caption{Counterfactual search on synthetic datasets. The counterfactuals obtained for the S surface and Swiss roll illustrate the role of diffusion distance (panels b and d) to take into account the connectivity of the data manifold as opposed to $L_{1}$ (panels a and b) or more generally other $L_{p}$ distances.}
    \label{fig:synthetic-all}
\end{figure}

We applied the CoDiCE framework on synthetic datasets to illustrate the effect of diffusion distance on geometrically structured data. Figure ~\ref{fig:synthetic-all} shows the "S" surface and Swiss roll shape partition in two classes. We took original Input point from class 0 and searched for counterfactual point minimising a $L_1$ distance (Figure ~\ref{fig:synthetic-all} (a), (c)), and diffusion distance (Figure ~\ref{fig:synthetic-all} (b), (d)). As indicated in the figure, the counterfactual obtained using $L_1$ distance crosses a low-density data region and ignores geometrical structure of the data, while counterfactuals found using diffusion distance respects the connectivity of the data manifold. 

\begin{figure}[!htbp]
    \centering
    \begin{subfigure}[b]{0.47\linewidth}
        \includegraphics[width=\linewidth]{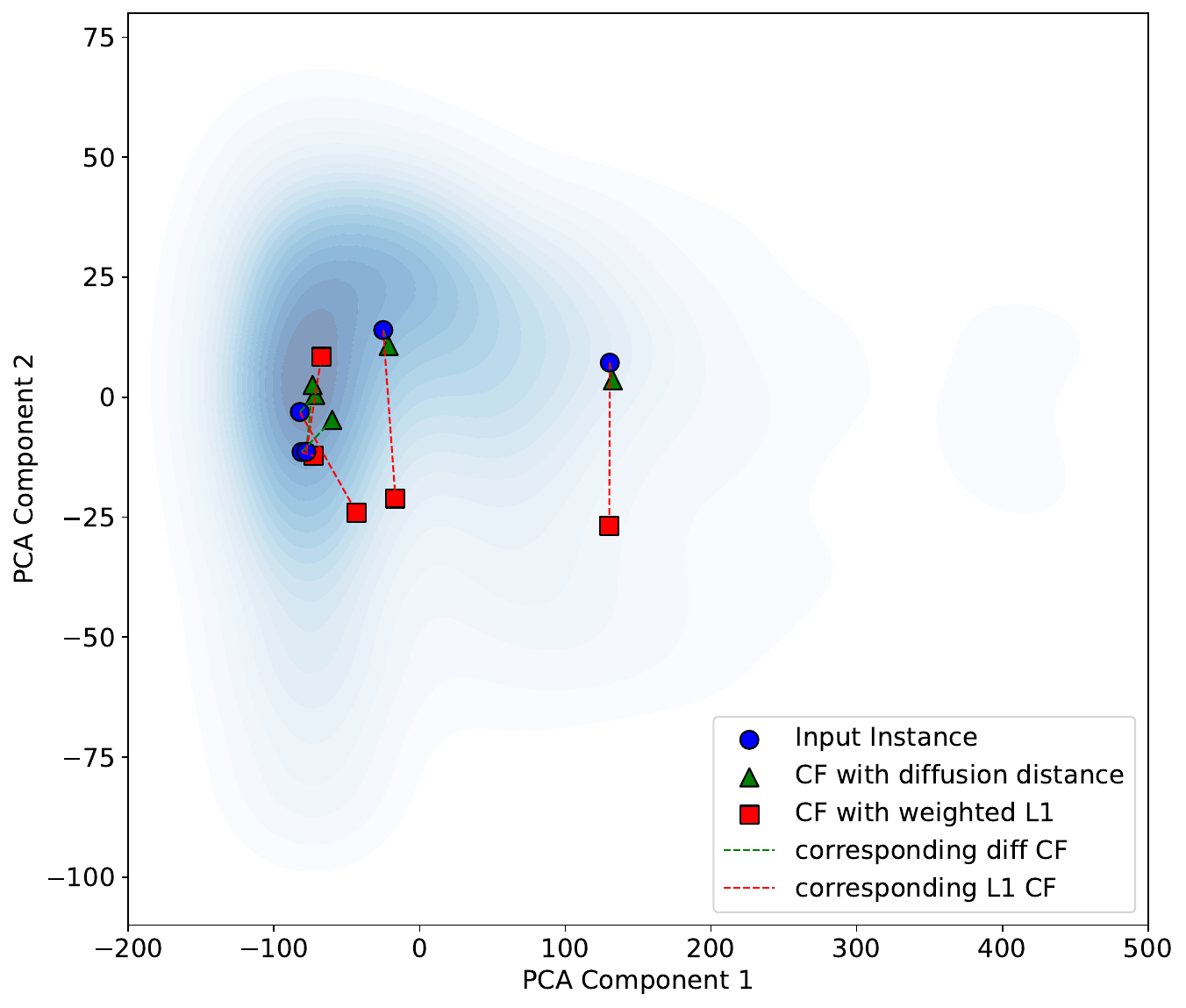}
        \caption{}
        \label{fig:pca-diab}
    \end{subfigure}
    \hfill % Optional spacing
    \begin{subfigure}[b]{0.47\linewidth}
        \includegraphics[width=\linewidth]{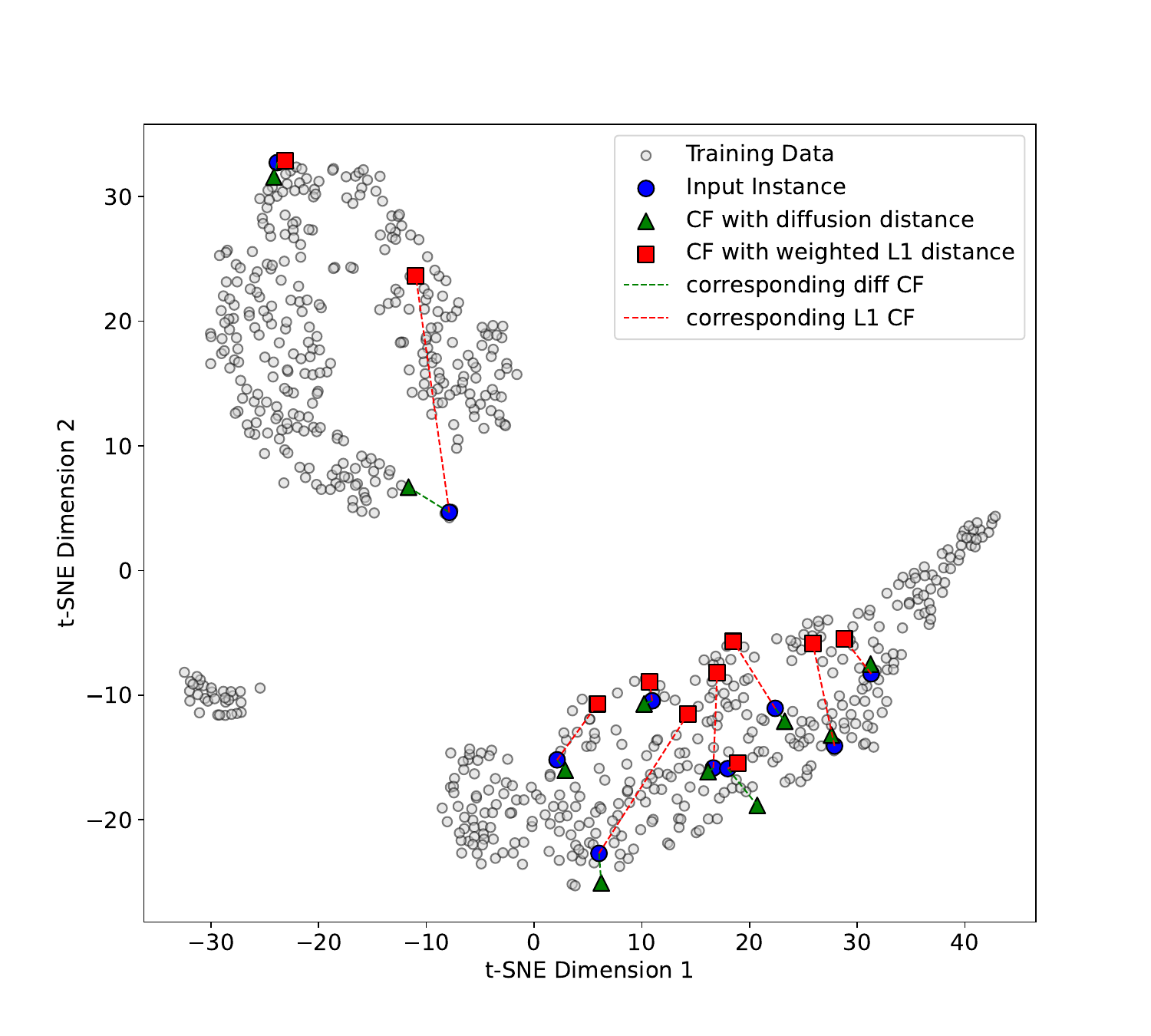}
        \caption{}
        \label{fig:tsne-diab}
    \end{subfigure}
    \caption{Counterfactual search on the Diabetes dataset projected onto PCA coordinates (a) and t-SNE (b). The plot has triplets of points connected by doted lines Input instance (blue circle) their respective Counterfactual point obtained with diffusion distance (green triangle) and Counterfactual point obtained with weighted $L_1$ distance (red square). }
    \label{fig:diabetes-diff}
\end{figure}

To illustrate the effect of diffusion distance on the counterfactual search for high-dimensional data, we also applied dimensionality reduction techniques such as PCA and t-SNE. For the trained Logistic Regression model and diabetes test set we searched for counterfactual explanations with weighted $L_1$ ($L_1$ norm with the inverse of the median absolute deviation that is commonly used in counterfactual methods as it originates from Wachter et al.\cite{wachter2017counterfactual}) as well as diffusion distance. We evaluated the search using 5 random samples as original inputs and visualize them together with their counterfactuals on PCA and 10 random samples visualized on t-SNE coordinates (Fig.~\ref{fig:diabetes-diff}). The counterfactuals obtained using diffusion distance point are significantly closer in both PCA and t-SNE coordinates than the corresponding obtained with L1 distance. This type of visualizations further suggest that diffusion distances are useful to capture the connectivity and clustering structure of the dataset.   

\subsection{Comparison of CoDiCE with other counterfactual methods on various datasets}
Beyond the visual illustrations, we set to compare the effect of diffusion distance and directional coherence against various popular conterfactual frameworks. In particular, we evaluated the following statistics: validity, diffusion distance, weighted L1 distance for continuous features, L0 distance for categorical features and directional coherence.

To make a fair comparison we adjusted the preprocessing of the data and our choice of model to be compatible with other frameworks. For all datasets we apply OneHotEncoding for categorical features and standardization for continuous features. After that we split the data into 80/20 train test split to train Logistic Regression model. In our experiments model is treated as a black-box and can be replaced with any other model. Finally, we generated counterfactual explanations for the first 100 instances from the same test set across all methods for every dataset.

\begin{table}[htbp]
\centering
\caption{Evaluation metrics comparison across different frameworks for datasets with continuous features. For distance and coherence metrics we report average and standard deviation over 100 samples. Validity is expressed in \%.}
\label{tab:cont_comparison}
\begin{tabular}{>{\raggedright\arraybackslash}p{2cm}ccccc}
\toprule
Dataset  & Metric& Validity $\uparrow$& Diffusion $\downarrow$& L1 continuous $\downarrow$& DCoherence $\uparrow$\\
\midrule
\multirow{7}{*}{Diabetes} 
& CoDiCE\textsubscript{diff}& \textbf{100\%} & \textbf{0.38 $\pm$ 0.22}& $1.11 \pm 0.53$& $0.64 \pm 0.15$\\
         & CoDiCE\textsubscript{$L_1$}& \textbf{100\%} & $0.72 \pm 0.45$& \textbf{0.29 $\pm$ 0.16}& $0.76 \pm 0.16$\\
         & DiCE  & 54\%& $1.62 \pm 0.73$& $1.10 \pm 0.34$& $0.68 \pm 0.13$\\
& FACE   & 70
\%& $1.64 \pm 0.67$& $1.08 \pm 0.36$& $0.72 \pm 0.14$\\
         & Prototypes& 26\%& $2.18 \pm 0.88$& $2.12 \pm 0.61$& \textbf{0.84 $\pm$ 0.07}\\
& GS& \textbf{100\%}& $0.67 \pm 0.31$& $0.39 \pm 0.19$& $0.57 \pm 0.12$\\
\midrule
\multirow{7}{*}{Breast Cancer}
& CoDiCE\textsubscript{diff}& 60\%& $2.87 \pm 1.21$& $2.18 \pm 0.39$& \textbf{0.78 $\pm$ 0.10}\\
& CoDiCE\textsubscript{$L_1$}& 60\%& $2.62 \pm 1.07$& $1.22 \pm 0.37$& $0.67 \pm 0.09$\\
         & DiCE  & 46\%& \textbf{2.13 $\pm$ 0.87}& \textbf{0.97 $\pm$ 0.28}& $0.72 \pm 0.08$\\
         & FACE   & 63\%& $2.91 \pm 1.08$& $0.98 \pm 0.38$& $0.74\pm 0.10$\\
         & Prototypes& 31\%& $4.72 \pm 0.45$& $2.22 \pm 0.37$& \textbf{0.79 $\pm$ 0.01}\\
& GS& \textbf{100\%}& $2.25 \pm 1.31$& $0.47 \pm 0.28$& $0.58 \pm 0.08$\\
\bottomrule
 & & & & & \\
\end{tabular}
\end{table}

To measure the effect of directional coherence isolated from that of the diffusion distance, we implemented two versions of CoDiCE framework: CoDiCE\textsubscript{diff} that uses diffusion distance for proximity and directional coherence CoDiCE\textsubscript{$L_1$} that uses a weighted $L_1$ proximity measure similarly to DiCE and directional coherence. Table~\ref{tab:cont_comparison} shows the comparison of methods for Diabetes and Breast Cancer datasets, both of which represent data with continuous features. 

We report that for the Diabetes dataset only CoDiCE\textsubscript{diff}, CoDiCE\textsubscript{$L_1$}, and GS obtained 100\% Validity, and hence all of the proposed counterfactual points for these methods did actually flip the outcome. It is important to note that the metrics are computed for valid points (points for which the desired outcome change is realized). As expected CoDiCE\textsubscript{diff} has the lowest diffusion distance ($0.38 \pm 0.22$) and both CoDiCE\textsubscript{$L_1$} has the lowest weighte L1 distance.  FACE showed to be more efficient method than Prototypes having higher Validity, which resulted in the highest directional coherence score, given that it was computed on a very few valid counterfactuals ($0.84 \pm 0.07$). Surprisingly, GS is comparable with CoDiCE\textsubscript{$L_1$} in terms of proximity scores, however directional coherence is the lowest.

For Breast Cancer dataset we found that it is quite difficult to find both directionally coherent and feasible points. The validity of all methods was low except for GS, which produces the least coherent counterfactuals. The highest performing methods in terms of directional coherence were Prototypes ($0.78 \pm 0.10$) and CoDiCE\textsubscript{diff} ($0.79 \pm 0.01$). However, Prototypes' validity only reached 31\%, and hence it discovered less than half the number of valid counterfactuals than CoDiCE methods found (which reached 60\% validity). Overall, as we will explicitly show in ablation experiments directional coherence is in a trade-off with diffusion distance (see Fig.~\ref{fig:abl-coh}) which resulted in the algorithm not having lowest diffusion distance for this setting (specific weights for each term in the cost function).

\begin{table}[htbp]
\centering
\caption{Evaluation metrics comparison across different frameworks for dataset with mixed-type of features (continuous + categorical). For distance and coherence metrics we report average and standard deviation over 100 samples. Validity is expressed in \%.}
\label{tab:mixed_comparison}
\begin{tabularx}{\textwidth}{>{\raggedright\arraybackslash}Xcccccc}
\toprule
Dataset  & Metric& Validity $\uparrow$& Diffusion $\downarrow$& $L_1$ cont $\downarrow$& $L_0$ cat $\downarrow$& DCoherence $\uparrow$\\
\midrule
\multirow{5}{*}{Adult}
& CoDiCE\textsubscript{diff}& \textbf{98\%}& \textbf{0.001 $\pm$ 0.004}& $3.9 \pm 1.4$& $0.2 \pm 0.1$& $0.84 \pm 0.09$\\
         & CoDiCE\textsubscript{$L_1$}& 92\%& $0.005 \pm 0.013$& $1.1 \pm 0.6$& $0.4 \pm 0.1$& $0.82 \pm 0.08$\\
         & DiCE  & 78\%& $0.005 \pm 0.011$& $1.2 \pm 0.6$& \textbf{0.1 $\pm$ 0.1}& \textbf{0.98 $\pm$ 0.02}\\
& FACE   & 82\%& $0.007 \pm 0.013$& \textbf{0.7 $\pm$ 0.3}& $0.5 \pm 0.2$& $0.81 \pm 0.11$\\
         & Prototypes& 19\%& $0.013 \pm 0.012$& $1.6 \pm 0.4$& $0.7 \pm 0.1$& $0.85 \pm 0.05$\\
\midrule
         \multirow{5}{*}{German}
         & CoDiCE\textsubscript{diff}& \textbf{100\%} & \textbf{4.3 $\pm$ 3.4}& $1.2 \pm 0.5$& \textbf{0.1 $\pm$ 0.1}& $0.93\pm0.04$\\
         & CoDiCE\textsubscript{$L_1$}& \textbf{100\%} & $6.4\pm3.8$& \textbf{0.7 $\pm$ 0.4}& \textbf{0.1 $\pm$ 0.1}& \textbf{0.94$\pm$0.04}\\
 & DiCE  & 49\%& $8.2 \pm 3.2$& $1.2 \pm 0.4$& $0.5 \pm 0.1$& $0.78 \pm 0.07$\\
 & FACE& 63\%& $7.6 \pm 2.4$& $1.0 \pm 0.4$& $0.5 \pm 0.1$& $0.79 \pm 0.07$\\
         & Prototypes& 34\%& $10.1 \pm 3.4$& $1.1 \pm 0.6$& $0.7 \pm 0.1$& $0.73 \pm 0.05$\\
\midrule
\multirow{5}{*}{Compas}
& CoDiCE\textsubscript{diff}& \textbf{100\%} & $0.03 \pm 0.05$& $5.2 \pm 1.9$& \textbf{0} & $0.92 \pm 0.08$\\
         & CoDiCE\textsubscript{$L_1$}& \textbf{100\%} & $0.03 \pm 0.05$& \textbf{0.8 $\pm$ 0.4}& \textbf{0} & \textbf{0.93 $\pm$ 0.07}\\
         & DiCE & 49\%& $0.03 \pm 0.04$& $1.0 \pm 0.6$& $0.4 \pm 0.2$& $0.83 \pm 0.11$\\
& FACE   & 18\%& $0.04 \pm 0.06$& $1.2 \pm 0.6$& $0.5 \pm 0.2$& $0.79 \pm 0.11$\\
         & Prototypes& 18\%& \textbf{0.01 $\pm$ 0.01}& $1.3 \pm 0.7$& $0.6 \pm 0.1$& $0.76 \pm 0.09$\\
\bottomrule
\end{tabularx}
\end{table}

Table~\ref{tab:mixed_comparison} shows the comparison of counterfactual methods on mixed-type datasets, such as Adult, German, Compas. Similarly as for the continuous case, counterfactual instances were generate for the first 100 instances of the fixed test set for the same Logistic Regression model. 

Notably, for all datasets CoDiCE\textsubscript{diff} and CoDiCE\textsubscript{$L_1$} results in the highest validity. For both Adult and German dataset CoDiCE\textsubscript{diff} has the lowest diffusion distance. However, we note that for Adult and Compas datasets the standard deviation of the diffusion distance is usually larger than its average. This indicates that diffusion distance distribution is significantly skewed for these datasets and the standard deviation might not fully capture the nature of its variability. Also for Compas the low validity for DiCE, FACE, and Prototypes resulted in the metrics for such methods to be evaluated from relatively few samples.

%and   while for Compas due to low Validity of other methods, which results in different number of points participating in comparison we argue that we can not compare the distances. 

Furthermore, mixed-type data brings the question of the weighting between the proximity metrics for continuous and categorical features. Depending on such weighting the method can results in cases in which a higher preference for changing categorical features results in a smaller difference in continuous features needed to flip the prediction. Hence, for mixed-type of data the pressure to minimize diffusion distance (which is exclusively computed for continuous features) is also a function of the weighting between continuous and categorical features proximity biases. 

As an additional metric, the running time for searching counterfactual explanations using these methods was measured. We report the running time for the most challenging dataset in terms of diffusion distance: Breast Cancer, which has 30 continuous features. The average speed of finding one counterfactual point on such a challenging dataset averaged 27 seconds for CoDiCE\textsubscript{diff} and 26 seconds for CoDiCE\textsubscript{$L_1$}. The running times are comparable since the diffusion distance is calculated for the entire dataset once. It is important to note that no code optimization efforts have been made thus far. For other methods, FACE requires, on average, 13 seconds to identify a counterfactual point, GS took 0.44 seconds, and the highly optimized DiCE took 0.11 seconds per counterfactual point. Although not the focus of our current investigation, we believe that code optimization efforts could significantly enhance running times, an aspect we aim to explore in forthcoming studies.   

\begin{table}[htbp]
\centering
\caption{Evaluation metrics  comparison across different frameworks for the Energy consumption dataset (regression problem). For distance and coherence metrics we report average and standard deviation over 100 samples. Validity is expressed in \%.}
\label{tab:energy_comparison}
\begin{tabular}{lccccc}
\toprule
Dataset  & Metric& Validity $\uparrow$& Diffusion $\downarrow$& $L_1$ continuous $\downarrow$& DCoherence $\uparrow$\\
\midrule
Energy & CoDiCE\textsubscript{diff} & 100\% & $0.005 \pm 0.03$ & $0.52 \pm 0.33$ & \textbf{0.67 $\pm$ 0.04}\\
         & CoDiCE\textsubscript{$L_1$}  & 100\% & \textbf{0.003 $\pm$ 0.02} & \textbf{0.41 $\pm$ 0.26} & $0.63 \pm 0.11$\\
         & DiCE   & 100\% & $1.64 \pm 2.21$ & $0.51 \pm 0.47$ & $0.62 \pm 0.11$\\
\bottomrule
\end{tabular}
\end{table}

The energy the consumption dataset was fitted with a genetic programming model (symbolic tree) trained with GP-GOMEA library \cite{virgolin2021improving}. It is a regression problem, which similarly to the logistic regression models for previous datasets, we treat as a black-box for counterfactual search. The target of interest was decreasing energy consumption by 5\%. We ran counterfactual search for 20 test instances, where for every instance we targeted an energy consumption decrease in the range [10\%, 5\%] of that predicted at the original input.
Among methods used in our comparison only DiCE supports a regression problem. The comparison with DiCE for this model is shown in Table~\ref{tab:energy_comparison}. 

\subsection{Ablation experiments}
To systematically explore the influence of each term within the CoDiCE objective function (5), we conducted ablation studies on the Diabetes dataset reported in Table~\ref{tab:metrics_comparison}. The ablations consist of setting the weights of all terms except the one under investigation to zero. Specifically, $\lambda_1$ corresponds to the weight assigned to the diffusion distance, $\lambda_2$ to the weight assigned to sparsity, and $\lambda_3$ to the weight assigned to directional coherence. 

\begin{table}[htbp]
\centering
\caption{Evaluation metrics under various ablations of diffusion, sparsity, and directional coherence terms for the Diabetes dataset.}
\label{tab:metrics_comparison}
\begin{tabularx}{\textwidth}{>{\raggedright\arraybackslash}Xcccccc}
\toprule
Dataset  & Inactive terms& Validity& Diffusion& $L_1$ continuous& Sparsity & DCoherence\\
\midrule
\multirow{3}{*}{Diabetes} & $\lambda_2, \lambda_3$ & $100\%$ & \textbf{1.49 $\pm$ 0.55}& \textbf{1.01 $\pm$ 0.48}& $1$& $0.56 \pm 0.13$\\
         & $\lambda_1, \lambda_3$ & $100\%$ & $2.38 \pm 0.86$& $1.92 \pm 0.43$& \textbf{0.85}& $0.56 \pm 0.13$\\
         & $\lambda_1, \lambda_2$  & $100\%$& $2.19 \pm 0.98$& $1.73 \pm 0.44$& $1$ & \textbf{0.82 $\pm$ 0.06}\\ 
\bottomrule
\end{tabularx}
\end{table}

\begin{figure}[!htbp]
    \centering
        \includegraphics[width=0.55\linewidth]{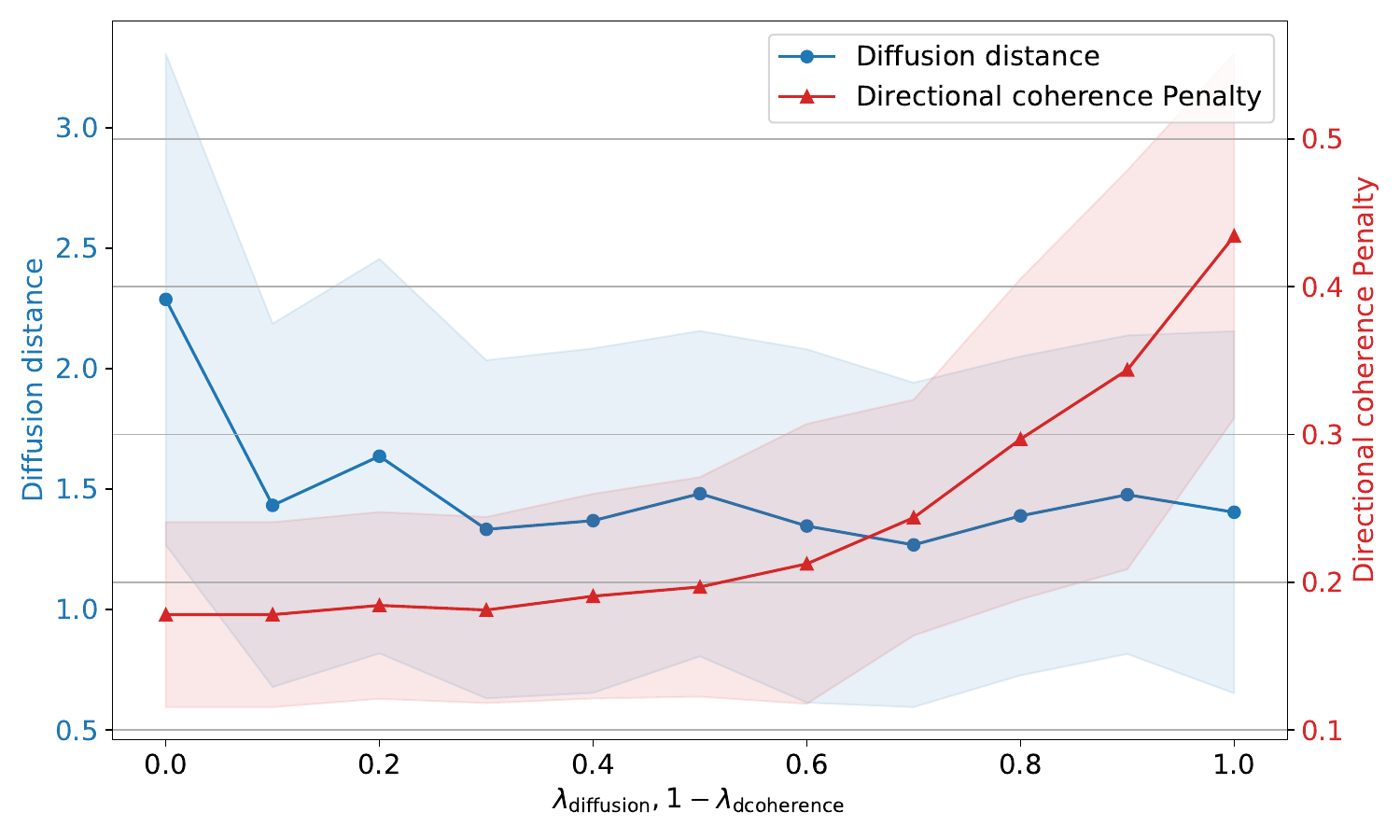}
        \caption{Trade-off between diffusion distance and directional coherence penalty is explored as the diffusion weight is increased.}    \label{fig:abl-coh}    
\end{figure}

Figure~\ref{fig:abl-coh} also illustrates the outcomes of trade-off experiments by evaluating the impact of varying the weights assigned to diffusion distance and directional coherence. Given a fixed sparsity weight $\lambda_2=0.5$, we varied the weight of diffusion distance $\lambda_1$ while setting the directional coherence weight to be $\lambda_3 = 1-\lambda_1$.  As the weight on diffusion distance increases from 0 to 1, effectively reducing the emphasis on directional coherence, we observe a notable trade-off. 

\section{Discussion}
This study proposes novel incorporations into an objective function of counterfactual search through the integration of diffusion distance and directional coherence. These are aimed to enhance the feasibility and alignment with intuitive expectations of a candidate counterfactual point. The introduction of diffusion distance as a component of the objective function ensures that the search for counterfactial points takes into account the underlying geometry of the data manifold. Additionally, diffusion distance is robust to noise which is particularly valuable in high-dimensional settings. The directional coherence term promotes that the counterfactual suggestion aligns with internal constrains or intuitions about how individual feature changes the model's outcome. After systematic experiments with various datasets (with both continuous and mixed-type data) we found that these terms work as intended and overall promote counterfactuals with shorter diffusion separation (being well connected or same cluster as the original input) and higher coherence (respecting marginal constraints). However, it is important to note that the inclusion of several terms simultaneously can result in a trade-off between both aims. The ablation experiments demonstrated notable trade-off between these two components, emphasizing the importance of a balanced approach to counterfactual explanation generation.
The search for explanation can be adjustable by the user identifying and weighing different biases, given that there are different usages for generating explanations or modus of construal \cite{keil_explanation_2006}, such as debugging the model behaviour or suggesting the end user to act on explanation.

\textbf{Future Work} could involve a deeper investigation into the role of diffusion distance in enhancing the robustness of counterfactual explanations. Given the importance of ensuring that generated counterfactuals closely mirror the underlying data distribution, diffusion distance may also contribute to making explanations more resistant to variations in data or model perturbations. It would be interesting to assess the impact of diffusion distance on the stability of counterfactual explanations under varying conditions of data noise and model uncertainty. 
While this study has highlighted the importance of balancing diffusion distance and directional coherence, the generation of diverse counterfactuals that equally satisfy these criteria presents a challenge. Addressing this challenge could involve the adoption of multi-objective optimization strategies, where diffusion distance and directional coherence are simultaneously optimized. That would enrich the set of available counterfactual explanations and allow to adapt towards user preferences, enhancing applicability. 

\section{Conclusion}
We explored the effect of integrating diffusion distance and directional coherence into the counterfactual explanation generation process. The proposed approach produces explanations that are more aligned with human intuition about transition from factual to counterfactual point and take into account the intrinsic geometry of the data. The findings highlight a crucial trade-off between these two factors, underlining the necessity for a balanced integration to optimize explanation quality. By formalising insights from human cognitive processes into AI explanation frameworks, the field is advancing towards AI systems that offer both more human-centric and intuitive explanations. 

\begin{credits}
\subsubsection{\ackname}
This research was supported by the Estonian Research Council Grants PRG1604, the European Union’s Horizon 2020 Research and Innovation Programme under Grant Agreement No. 952060 (Trust AI), by the Estonian Centre of Excellence in Artificial Intelligence (EXAI), by the Estonian Ministry of Education and Research.

\subsubsection{\discintname}
The authors have no competing interests to declare that are
relevant to the content of this article.
\end{credits}

\section*{A Appendix}
To ensure reproducible research, the code for CoDiCE with all experiments is released in the public repository:

\begin{verbatim}
    https://github.com/anitera/CoDiCE
\end{verbatim}

\subsection*{Diffusion distance with Self-tuning kernel}
Our implementation incorporates a self-tuning kernel within the diffusion distance framework \cite{zelnik2004self}, adjusting dynamically to the variance in the data. The main difference of self-tuning kernel approach is the kernel used for local scaling:

\begin{equation}
K(x, y) = \exp\left(-\frac{\|x - y\|^2}{\sigma_x \sigma_y}\right) \, .
\end{equation}

\noindent Here, $K(x, y)$ denotes the kernel similarity between points $x$ and $y$, while $\sigma_x$ and $\sigma_y$ are the local scaling parameters for $x$ and $y$, respectively. These parameters are typically determined based on the distances to their $k$-th nearest neighbors, allowing the kernel to adaptively modulate its influence across different data densities.

The transition matrix $P$, pivotal for the diffusion process, is derived from the kernel matrix $K$ through normalization:
\begin{equation}
P = D^{-1}K
\end{equation}

where $D$ is the diagonal degree matrix of kernel similarity $D_{ii} = \sum_j K_{ij}$. This step converts the kernel similarities into transition probabilities, facilitating the computation of diffusion distances.

 For the exact implementation details of the method see Algorithm~\ref{alg:stdm}.

\begin{algorithm}
\caption{Self-Tuning Diffusion Maps (STDM)}\label{alg:stdm}
\begin{algorithmic}[1]
\Require Data matrix $X \in \mathbb{R}^{n \times d}$, number of neighbors $k$, diffusion time scale $\alpha$
\Ensure Diffusion map embeddings $DMap$, eigenvalues $\Lambda$, eigenvectors $V$
\State Initialize $STDiffusionMap$ object with $k$, $\alpha$
\State $K \gets$ construct\_affinity\_matrix$(X, k)$
\State $L \gets$ construct\_transition\_matrix$(K, \alpha)$
\State $DMap, V, \Lambda \gets$ make\_diffusion\_coords$(L)$
\Function{construct\_affinity\_matrix}{$X, k$}
    \State Compute $k$-nearest neighbors for $X$
    \State Calculate local scales for each data point
    \State Form affinity matrix $K$ using self-tuning kernel
    \State \Return $K$
\EndFunction
\Function{construct\_transition\_matrix}{$K, \alpha$}
    \State Calculate right normalization vector $q$
    \State Normalize $K$ to form transition matrix $P$
    \State Derive Laplacian $L$ from $P$
    \State \Return $L$
\EndFunction
\Function{make\_diffusion\_coords}{$L$}
    \State Compute eigenvalues and eigenvectors of $L$
    \State Select top eigenvalues and corresponding eigenvectors
    \State Calculate diffusion map embeddings $DMap$
    \State \Return $DMap, V, \Lambda$
\EndFunction
\end{algorithmic}
\end{algorithm}

\begin{algorithm}
\caption{Compute Directional Coherence Score}\label{alg:dircoh}
\begin{algorithmic}[1]
\Require Input instance $x$, Counterfactual instance $x'$, Required label $y'$, Model $M$, List of features $F$
\Ensure Directional coherence penalty $dir\_coherence$, Uncoherent suggestions $S$
\State Initialize $marginal\_signs \gets \{\}$ \Comment{Dictionary with the marginal direction of prediction change for each feature}
\State $N \gets \text{Length}(F)$
\For{\textbf{each} $f$ in $F$} \Comment{Iterate over all features}
    \State $marginal\_signs[f] \gets$ \Call{Marginal\_Pred\_Signs}{$x$, $x'$, $f$, $y'$, $M$} 
\EndFor

\State $dir\_coherence \gets \frac{1}{N} \sum_{f \in F} \left[ marginal\_signs[f] \neq -1 \right]$ \Comment{Calculate directional coherence score as ratio of coherent features}
\State $S \gets \{f \mid f \in F, marginal\_signs[f] = -1\}$ \Comment{Features not changing as expected}
\State \Return $dir\_coherence$, $S$

\Function{Marginal\_Pred\_Signs}{$x$, $x'$, $f$, $y'$, $M$}
    \State $control \gets$ copy($x$)
    \State $control[f].value \gets x'[f].value$ \Comment{Change only the current feature value}
    \If{$M.type$ == "classification"} 
        \State $original\_pred \gets M.predict\_proba\_instance(x)$ \Comment{If model gives access to probability prediction instead of thresholded value, it gives more precise result}
        \State $control\_pred \gets M.predict\_proba\_instance(control)$
        \State $prob\_sign \gets$ SIGN($control\_pred[y'] - original\_pred[y']$)
    \Else
        \State $original\_pred \gets M.predict\_instance(x)$
        \State $control\_pred \gets M.predict\_instance(control)$
        \State $prob\_sign \gets$ SIGN($control\_pred - original\_pred$)
    \EndIf
    \State \Return $prob\_sign$ \Comment{Sign of prediction change due to the feature's alteration}
\EndFunction
\end{algorithmic}
\end{algorithm}

\begin{algorithm}
\caption{Genetic Algorithm for Counterfactual Generation}
\label{alg:ga_counterfactual}
\begin{algorithmic}[1] % Enables line numbering
\Require $model, original\_instance, desired\_output, population\_size, max\_iterations$
\Ensure $counterfactuals$
\State Initialize population with $population\_size$ members for $original\_instance$
\State Evaluate fitness of each member in population using $model$ and $desired\_output$
\For{$iteration = 1$ to $max\_iterations$}
    \State Select top half of population based on fitness scores
    \State Generate offspring through crossover and mutation operations
    \State Evaluate fitness of new members
    \State Select $population\_size$ members for the next generation
    \If{convergence criteria met \textbf{or}  $iteration = max\_iterations$}
        \State Extract counterfactuals meeting $desired\_output$
        \State \textbf{break}
    \EndIf
\EndFor
\Return $counterfactuals$
\end{algorithmic}
\end{algorithm}

\subsection*{Evaluation metrics for counterfactual explanations}
\textbf{Validity} measures the proportion of generated counterfactuals that successfully achieve the desired outcome when applied to the model and defined as:
\begin{equation}
\text{Validity} = \frac{\text{Number of successful counterfactuals}}{\text{Total number of counterfactuals generated}}*100\%
\end{equation}

In evaluating counterfactual \textbf{proximity} to the original instance, it is standard practice to use the Weighted L1 distance for continuous features, as suggested by Wachter \cite{wachter2017counterfactual}, and the $L_0$ norm for categorical features \cite{guidotti_counterfactual_2022,karimi2021algorithmic}. While less common, some approaches incorporate the Mahalanobis distance \cite{kanamori_dace_2020} to assess the typicality of the counterfactual within the desired class distribution. For assessing continuous features, we use Weighted L1 Distance to maintain consistency with prior research, but we also add Diffusion distance. As compared to  Mahalanobis distance it is offering additional insights by considering the connectivity along the entire path from the factual point. 

\textbf{The Weighted L1} is defined by adjusting the L1 norm with the inverse of the median absolute deviation (MAD) for each feature. The formula is given by:
\begin{equation}
L1_{\text{Wachter}}(x, x') = \sum_{i=1}^{M} \left( \frac{|x_i - x'_i|}{\text{MAD}_i} \right)
\end{equation}
where $M$ is total number of points, $\text{MAD}_i$ is the median absolute deviation of the $i$-th feature across the dataset, and $x_i$ and $x'_i$ are the values of the $i$-th feature in the original and counterfactual instances, respectively.

\textbf{L0 Categorical} counts the number altered features and defined as:
\begin{equation}
L0(x, x') = \|\{i \mid x_i \neq x'_i\}\|_0
\end{equation}
indicating the count of non-zero differences between corresponding features of $x$ and $x'$.

\textbf{The Diffusion Distance} captures the proximity of the counterfactual to the original instance, taking into account the intrinsic geometry of the data manifold. It is calculated as:
\begin{equation}
D_{\text{diff}}(x, x') = \sqrt{\sum_{i=1}^{M} \frac{(p_t(x, i) - p_t(x', i))^2}{\phi_0(i)}}
\end{equation}
where $M$ is number of instances, $p_t(x, i)$ denotes the transition probability from point $x$ to $i$ in $t$ steps, and $\phi_0(i)$ represents the stationary distribution.

\textbf{Directional Coherence} assesses the consistency between the prediction changes of counterfactual features (jointly) and their per feature marginal directions with respect to the model prediction. Given an original instance as a vector $x=(x_1, x_2,\dots, x_n) \in \mathcal{X} \subseteq \mathbb{R}^n$ and a counterfactual instance $x^\ast=(x^\ast_1, x^\ast_2,\dots, x^\ast_n) \in \mathcal{X} \subseteq \mathbb{R}^n$ that brings the desired outcome. The goal is to evaluate the coherence of the transition from $x$ to $x^\ast$ in achieving a specified outcome label $y$ with a set of expected marginal transitions $$\{x_{i} \rightarrow x'_{i} \,\, | \,\, f(y|x_{1},x_{2},\ldots,x'_{i},\ldots,x_{n}) \geq f(y|x_{1},x_{2},\ldots,x_{i},\ldots,x_{n}) \, , 1 \leq i \leq n \}.$$  

Then, the Directional Coherence score counts the excess of features which have aligned marginal ($'$) and joint ($\ast$) directions to increase the model's prediction probability towards the desired outcome y:

\begin{equation}
dcoherence = \frac{1}{n}\sum_{i=1}^{n}\sgn\left(\left(x^\ast_i-x_i\right)\left(x'_i-x_i\right)\right) \, .
\end{equation}
%\bibliographystyle{splncs04}
%\bibliography{references}
%
\bibliographystyle{splncs04}
%\bibliography{references}

\end{document}